\documentclass[11pt]{article}

\usepackage[margin=1in]{geometry}
\usepackage{times}

\usepackage{amsmath,amsfonts,bm}

\def\eqref#1{equation~\ref{#1}}

\def\1{\bm{1}}

\DeclareMathAlphabet{\mathsfit}{\encodingdefault}{\sfdefault}{m}{sl}
\SetMathAlphabet{\mathsfit}{bold}{\encodingdefault}{\sfdefault}{bx}{n}

\usepackage{graphicx}
\usepackage{subcaption}
\usepackage{booktabs}
\usepackage{multirow}
\usepackage{xcolor}
\usepackage{amssymb}
\usepackage{amsmath}
\usepackage[numbers,sort&compress]{natbib}

\usepackage{hyperref}
\usepackage{url}
\newcommand{\modelname}{PIT-GCL}

\newcommand{\ms}[2]{$#1_{\scriptscriptstyle\pm#2}$}

\newcommand{\msb}[2]{
    $\mathbf{#1}_{\scriptscriptstyle\pm\mathbf{#2}}$
}

\newcommand{\PHGAIN}{$0.0157$}

\title{
PIT-GCL: Protein Interaction using Topological Graph Contrastive Learning
}

\author{
Jae Won Choi$^{1}$ \and
Ryoonki Hong$^{1}$ \and
Alan Liang$^{1}$ \and
Manjula Adiveppa Wader$^{1}$ \and
Bingsong Zeng$^{1}$ \and
Peiyang Tang$^{2}$ \and
Longwei Liu$^{2}$ \and
Ruishan Liu$^{1}$\\
\\
\small $^{1}$Department of Computer Science, University of Southern California \\
\small $^{2}$Department of Biomedical Engineering, University of Southern California \\
\small \texttt{jwchoi@usc.edu}
}

\date{}

\begin{document}

\maketitle

\begin{abstract}
Protein binding prediction is central to target identification, therapeutic binder design, and large scale screening, yet remains challenging because binding depends on sequence, three dimensional geometry, and global structural organization. Recent folding models such as AlphaFold3 and Boltz-2 have substantially improved structure prediction, but their confidence outputs (pLDDT, pTM, ipTM) are not specifically designed for binary binding prediction, and dedicated structure aware predictors often require bound complex structures that are unavailable at screening scale. We introduce \modelname{}, a dual tower structure aware framework that encodes each protein independently from its amino acid sequence, C$\alpha$ point cloud, and a global persistent homology descriptor. Each tower combines residue ESM-2 embeddings with a topological summary computed from the $H_0$ and $H_1$ persistence landscapes of a Vietoris-Rips filtration, and processes the resulting tokens with a structure aware Transformer in which pairwise C$\alpha$ distances enter as a learned attention bias. A bidirectional cross attention module then performs latent space soft docking between the two per-protein representations, and the model is trained with a combined binary cross entropy and NT-Xent contrastive objective. On three binary interaction prediction benchmarks, general PPI on PPIRef, TCR-pMHC binding on STAG, and whole chain pairs on PPB-Affinity, \modelname{} outperforms representative sequence based, structure aware, and task specific baselines on general PPI under our evaluation, and is the only method above chance on PPB-Affinity; on TCR-pMHC it leads at a fixed decision threshold but is outranked by a task specific sequence model. Because each protein is encoded independently in the first phase, its representation can be precomputed and reused across candidate pairs, which is convenient for large scale screening.
\end{abstract}


\section{Introduction}
\label{sec:intro}

Predicting whether two proteins physically interact underpins target
identification, therapeutic binder design and large scale screening, and is hard
because binding depends on sequence, 3D geometry and global structural
organization. Folding models such as AlphaFold-Multimer~\citep{Evans2022AFMultimer},
AlphaFold3~\citep{Abramson2024AlphaFold3} and Boltz-2~\citep{Passaro2025Boltz2}
predict complex structure rather than a calibrated binary score, and all of them
co-fold the complex at inference. At minutes per pair, a screen over $10{,}000$
pairs costs on the order of $10^3$ GPU-hours, while \modelname{} scores the same
set in $17$ minutes on one GPU, of which the part that scales with the number of
candidates is under a minute and the rest is a one-off descriptor computation a
second target reuses (Appendix~\ref{app:runtime}).

We propose \modelname{}, a general purpose protein binding prediction framework that predicts binding without requiring the bound complex structure as input. \modelname{} uses a topology aware dual tower design: each protein is encoded independently from its sequence, C$\alpha$ point cloud, and persistent homology descriptor, and pairwise interactions are modeled only after protein level representations are constructed. This formulation preserves protein specific information before interaction level reasoning and allows per-protein embeddings to be reused across candidate pairs in large scale screening.

We evaluate \modelname{} on three binary interaction prediction tasks and they
do not give the same answer, which we report as such. On general PPI
(PPIRef~\citep{PPIformer2024ICLR}) it leads every baseline we rank on all five
metrics. On TCR-pMHC (STAG~\citep{STAG2025}) it leads at a fixed decision
threshold but is outranked by a task-specific sequence model, so we make no
state-of-the-art claim there. On PPB-Affinity~\citep{PPBAffinity2024}, which
applies the same label construction to whole chains rather than interface
regions, it is the only method above chance, though it recovers only part of the
task.

Our contributions are: a topology-aware dual-tower framework that encodes each
protein independently and predicts binding through cross-protein attention,
needing no bound complex; an injection of a global persistent-homology vector at
the residue level before structure-aware attention, whose contribution we
quantify with parameter-matched and shuffled controls; and an evaluation on
three benchmarks with every baseline retrained on the benchmark it is scored on,
separating where the model leads from where it does not. Code is uploaded as
supplementary material.

\section{Related Work}
\label{sec:related}

Methods for protein-protein binding prediction differ in the input modality they
consume - sequence, individual structures, or a co-folded complex - and in what
each side of a pair actually is: general PPI benchmarks such as
PPIRef~\citep{PPIformer2024ICLR} present the extracted interface region of each
chain, PPB-Affinity~\citep{PPBAffinity2024} presents whole chains, and
specialized settings such as TCR-pMHC fix the interaction family instead. We
evaluate all three and organize prior work by input modality. \modelname{} sits
between the sequence-only and co-folding families: it takes structure from
individual chains without requiring a co-folded complex, and produces
per-protein representations that can be cached and reused.

\paragraph{Protein encoders.}
Sequence encoders range from substitution-matrix embeddings such as
BLOSUM~\citep{Henikoff1992BLOSUM}, still common in TCR-pMHC predictors, to
protein language models such as ProtTrans~\citep{Elnaggar2022ProtTrans} and
ESM-2~\citep{ESM22023}. Structure encoders operate on residue point clouds or
graphs, with equivariant designs including the Geometric Vector
Perceptron~\citep{Jing2021GVP}, $E(n)$-equivariant graph
networks~\citep{Satorras2021EGNN} and the Equiformer~\citep{Liao2023Equiformer}
used inside PPIFormer~\citep{PPIformer2024ICLR}. \modelname{} pairs a frozen
ESM-2 encoder with a lighter geometric strategy: a structure-aware Transformer
over the C$\alpha$ point cloud plus a global topological summary of it.

\paragraph{Topological descriptors for biomolecules.}
Persistent homology has been applied to biomolecular property prediction since
TopologyNet~\citep{CangWei2017TopologyNet}, which feeds persistence barcodes of
protein-ligand complexes to a convolutional network, and its element-specific
successor~\citep{CangWei2018ElementSpecific}, which partitions the point cloud
by atom type so that each barcode carries chemical identity. For interactions
specifically, persistent spectral methods reach binding affinity through the
eigenvalues of Laplacians built on the filtration rather than through barcodes
alone~\citep{WeeXia2022PerSpect}. Those lines share a design that ours does not:
the descriptor is computed on the \emph{complex} and enriched with atom or
residue attributes, which makes it interface-specific and correspondingly
dependent on having the complex. \modelname{} computes a single
attribute-free landscape per protein from C$\alpha$ positions alone, which is
weaker per descriptor but is what allows the two towers to be encoded and cached
independently. The name is also worth separating from
TopoGCL~\citep{Chen2024TopoGCL}, which contrasts augmented views of a graph;
our contrastive term aligns two protein towers and involves no graph views.

\subsection{Protein-Protein Interaction Prediction}

\paragraph{Sequence-based methods.}
Sequence-based predictors learn directly from a pair of amino acid sequences.
DeepDTA~\citep{DeepDTA2018} stacks 1D convolutions with global max pooling and a
fully connected head, and PIPR~\citep{PIPR2019} uses a Siamese residual
recurrent convolutional network over amino acid embeddings. Both are easy to
train at scale but ignore three-dimensional geometry, which limits their ability
to separate sequence-similar pairs that behave differently in 3D.

\paragraph{Structure-aware methods.}
D-SCRIPT~\citep{DSCRIPT2021} embeds each sequence with a pretrained language
model and pools a predicted inter-chain contact map into a binding probability;
PPIFormer~\citep{PPIformer2024ICLR} represents each interface as an
SE(3)-equivariant point cloud, pretrains an Equiformer with masked modeling on
PPIRef, and scores pairs by average log-likelihood. Predictors that operate on
bound complexes or co-folded outputs~\citep{Gainza2020MaSIF,Reau2023DeepRank,Evans2022AFMultimer}
are excluded from our screening-style evaluation, which assumes only individual
structures. A practical caveat of this family is calibration: PPIFormer emits an
unnormalized log-likelihood rather than a probability, so a fixed $0.5$ threshold
penalises it disproportionately (Appendix~\ref{app:threshold}).

\subsection{TCR-pMHC Binding Prediction}

Here one partner is a T-cell receptor, characterized primarily by its
CDR3$\alpha$ and CDR3$\beta$ loops, and the other a peptide presented on an MHC.
Most methods are sequence-level: 1D CNNs over BLOSUM-encoded CDR3 and peptide
sequences~\citep{NetTCR22021}, a Random Forest with an MHC-aware
variant~\citep{epiTCR2023}, interpretable attention~\citep{TEPCAM2024}, joint
attention over the TCR and a SMILES-encoded peptide~\citep{Titan2021}, and
contrastive objectives~\citep{TRAP2025,ConTCR2025,ProTCR2026}. None reasons
directly about the 3D arrangement of the CDR loops. On the structural side,
STAG~\citep{STAG2025} builds a graph from the bound complex, which captures
interface geometry but requires a resolved or modelled complex for every
candidate pair. \modelname{} differs from
both: it encodes each protein from its individual structure and fuses the pair
only through cross-attention, removing the dependence on a pre-existing complex
at inference time.

\section{Method}
\label{sec:method}
\modelname{} encodes each protein of a pair independently and models their
interaction only afterwards, through cross-protein attention. The decoupled
design preserves protein-specific sequence, geometric and topological
information before fusion and yields per-protein representations that can be
cached and reused at inference.

The encoder follows the graph-Transformer paradigm, in which each node is a
token and relational structure enters as an additive bias on the attention
logits rather than as message passing along
edges~\citep{Ying2021Graphormer,Kreuzer2021SAN,Park2022GRPE}; for protein
structures that bias has been instantiated with residue distances as in the pair
representation of the Evoformer~\citep{Jumper2021AlphaFold}, and we do the same
at the residue level with pairwise C$\alpha$ distances. The global descriptor
comes from persistent homology, which tracks topological invariants across a
filtration, is invariant under rigid motions and is provably stable under small
perturbations~\citep{CohenSteiner2007Stability}; we vectorise it as persistence
landscapes~\citep{PersistenceLandscape2020,bubenik2015statistical}, with a
primer in Appendix~\ref{app:tda_primer}.

\paragraph{Phase 1: Per-Protein Embedding.}

\begin{figure}[t]
    \centering
    \includegraphics[width=\linewidth]{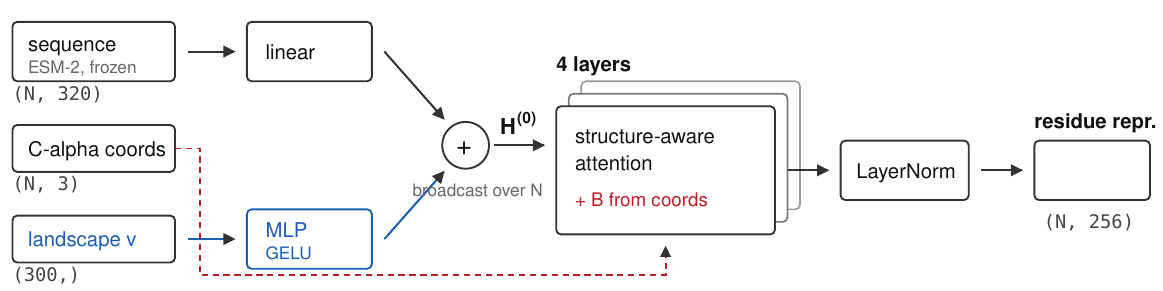}
    \caption{The per-protein encoder. A frozen ESM-2 embedding is projected to
    the model width, a two-layer MLP maps the persistence landscape to the same
    width, and the two are added so every residue token carries the protein's
    global topology; four structure-aware layers then attend over the residues
    with a bias built from the C$\alpha$ coordinates
    (Figure~\ref{fig:struct_bias}). How the landscape itself is built is shown
    in Figure~\ref{fig:Method_Topo}.}
    \label{fig:Method_Overview}
\end{figure}

\modelname{} applies the same encoder to each protein of a pair $(P_A, P_B)$
independently (Figure~\ref{fig:Method_Overview}). For a protein $P$ with $N$ residues we extract three modalities. First, the C$\alpha$ coordinates $X \in \mathbb{R}^{N \times 3}$, which provide residue-level 3D geometric information. Second, we obtain residue-wise sequence embeddings from a pretrained ESM-2~\citep{ESM22023} language model, $S \in \mathbb{R}^{N \times d_{\mathrm{esm}}}$, which capture evolutionary and semantic information from the amino acid sequence. Third, we compute a persistent-homology-based topological descriptor $v \in \mathbb{R}^{d_{\mathrm{topo}}}$ by constructing a Vietoris-Rips filtration~\citep{Vietoris1927,EdelsbrunnerHarer2010} on the C$\alpha$ point cloud. Since computing $H_k$ requires simplices of dimension $k+1$ (Appendix~\ref{app:tda_primer}), the complex is expanded to dimension $2$. The resulting $H_0$ and $H_1$ persistence diagrams are summarized as persistence landscapes~\citep{bubenik2015statistical} and concatenated into a single vector that captures the global topological organization of the protein.

We project the residue-wise ESM-2 embeddings into the model hidden dimension via a linear map, $H_{\mathrm{seq}} = S W_{\mathrm{esm}} \in \mathbb{R}^{N \times d_{\mathrm{model}}}$, and map the global topological descriptor to the same dimension using a two-layer MLP with a GELU activation, $h_{\mathrm{topo}} = \mathrm{MLP}_{\mathrm{topo}}(v) \in \mathbb{R}^{d_{\mathrm{model}}}$. We then inject the global topological signal into every residue representation by broadcasting and adding it to the residue-wise sequence embedding,
\[
H^{(0)} = H_{\mathrm{seq}} + \mathbf{1}_N\, h_{\mathrm{topo}}^\top \in \mathbb{R}^{N \times d_{\mathrm{model}}},
\]
so that each residue token is conditioned on a single shared summary of the protein's topology before any structural reasoning takes place.

To incorporate 3D geometric information, $H^{(0)}$ is processed by a stack of $L$ structure-aware Transformer layers in which residues serve as tokens and pairwise C$\alpha$ distances enter as additive attention biases. For residues $i$ and $j$ we compute $d_{ij} = \|X_i - X_j\|_2$, encode the resulting distance matrix with $16$ radial basis functions on the interval $[0,\,20]\,\text{\AA}$, and project the RBF features through a linear layer to obtain a per-head attention bias $B_{\mathrm{struct}} \in \mathbb{R}^{H \times N \times N}$ (Figure~\ref{fig:struct_bias}, Appendix~\ref{app:ablation_full}). The structure-aware attention operation is then defined, per head, as
\[
\mathrm{Attn}(Q,K,V)
=
\mathrm{softmax}\!\left(
\frac{QK^\top}{\sqrt{d_h}} + B_{\mathrm{struct}}
\right)V,
\]
where $d_h = d_{\mathrm{model}}/H$; head outputs are concatenated and combined by an output projection. We use multi-head attention with $H$ heads throughout and omit head indices for clarity. Each block follows a pre-norm architecture with residual connections and a position-wise feed-forward network with GELU activation. After $L$ layers, the encoder produces residue-level representations for the two proteins, $H_A \in \mathbb{R}^{N_A \times d_{\mathrm{model}}}$ and $H_B \in \mathbb{R}^{N_B \times d_{\mathrm{model}}}$.

\paragraph{Phase 2: Interaction Prediction.}
Given the per-protein representations $H_A$ and $H_B$, \modelname{} models their interaction with a bidirectional cross-attention module. This step can be viewed as a latent-space soft-docking operation: rather than explicitly docking the two structures in 3D, the model allows residues in one protein to attend to residues in the other.

Query, key and value projections are protein-specific rather than tied across
the two towers, $Q_A = H_A W_Q^A$, $K_A = H_A W_K^A$, $V_A = H_A W_V^A$ and
likewise $W_\cdot^B$ for $B$, which lets each tower specialize according to its
role in the interaction. For $A$ attending to $B$,
\[
H_{A \leftarrow B}
=
\mathrm{softmax}\!\left(
\frac{Q_A K_B^\top}{\sqrt{d_h}}
\right)V_B,
\qquad
H_A' = \mathrm{LayerNorm}(H_A + H_{A \leftarrow B}),
\]
and symmetrically for $B$ attending to $A$. Both directions use multi-head
attention with an output linear projection.

To form pair-level representations, we mean-pool the updated residue embeddings, $z_A = \mathrm{MeanPool}(H_A')$ and $z_B = \mathrm{MeanPool}(H_B')$, concatenate them into $Z = [z_A \,\|\, z_B] \in \mathbb{R}^{2 d_{\mathrm{model}}}$ and pass the result through a prediction head,
\[
\hat{y} = \sigma\!\left(\mathrm{MLP}\big(\mathrm{LayerNorm}(Z)\big)\right),
\]
where $\hat{y} \in (0,1)$ is the predicted probability that the two proteins interact.

\paragraph{Training Objective.}
We combine a binary cross-entropy term
$\mathcal{L}_{\mathrm{BCE}} = -[\, y \log \hat{y} + (1-y)\log(1-\hat{y})\,]$
on the predicted probability with a contrastive term applied directly to the
pooled embeddings $z_A$ and $z_B$ produced after cross-attention, without any
additional projection head. Within a mini-batch of $B$ pairs we
$\ell_2$-normalize the embeddings and form a similarity matrix between the $A$-
and $B$-side vectors, $\mathrm{sim}_{ij} = \langle z_A^{(i)},\, z_B^{(j)} \rangle / \tau$ with $\tau$ a temperature. For each interacting pair (label $y_i = 1$), the matched index $i$ is the only positive on its row and column; all other indices in the mini-batch act as in-batch negatives. We use a symmetric NT-Xent
objective~\citep{Sohn2016NPair,Oord2018CPC,Chen2020SimCLR} averaged over the
positive pairs of the batch in both directions, and train on
$\mathcal{L} = \alpha\,\mathcal{L}_{\mathrm{NT\text{-}Xent}} +
\beta\,\mathcal{L}_{\mathrm{BCE}}$ with mixing weights $\alpha$ and $\beta$.
The contrastive term aligns each protein-$A$ embedding with its matched partner
relative to in-batch alternatives, complementing the BCE term at the
representation level; Appendix~\ref{app:implementation_details} writes it out.

\paragraph{Inference and Scalability.}
Phase 1 depends on one protein only, so its residue representations can be
precomputed and cached and only Phase 2 is recomputed per candidate pair. With
individual structures rather than complexes as input, this is what makes the
design usable at screening scale (Appendix~\ref{app:runtime}).

\section{Experiments}
\label{sec:experiments}

\subsection{Evaluation Protocol}

\paragraph{Datasets.}
We evaluate on three binary interaction tasks, each under its own benchmark's
protocol rather than a partition of our choosing. For general PPI we use
PPIRef~\citep{PPIformer2024ICLR}, taking the authors' filtered and
iDist-clustered fold - their near-duplicate threshold of $0.04$ reduces
$349{,}685$ filtered interfaces to $51{,}755$ cluster representatives, and this
is the fold their release prescribes - which we partition $80/10/10$ per seed,
with their re-pairing protocol for negatives. For TCR-pMHC we use the STAG
benchmark's own train/test split~\citep{STAG2025} unchanged, carving a
validation set from its training half because the benchmark provides none. We
additionally evaluate on
PPB-Affinity~\citep{PPBAffinity2024}, a curated set of complexes with measured
binding affinities. Both PPIRef and PPB-Affinity supply positive pairs only, so
on both we build negatives the same way, by re-pairing a chain with one drawn
from a different complex; the measured affinities are not thresholded and do not
enter the label. What differs is the input: a PPIRef row is the $6\,\text{\AA}$
interface region of each chain, a median of $24$ residues per tower, while a
PPB-Affinity row is the whole chain, a median of $199$. PPB-Affinity therefore
runs the same architecture and the same label construction on a roughly
eight-fold larger input. It uses an $80/10/10$ partition of
$10{,}484$ balanced pairs ($8{,}414/1{,}036/1{,}034$). Split statistics and the
provenance of the coordinates - which are the chains of resolved complexes on
all three benchmarks - are in Appendix~\ref{app:Dataset_Details}, and the
screening throughput the design is motivated by is measured in
Appendix~\ref{app:runtime}.

\paragraph{Baselines.}
On PPIRef we compare against DeepDTA~\citep{DeepDTA2018}, PIPR~\citep{PIPR2019},
D-SCRIPT~\citep{DSCRIPT2021}, PPIFormer~\citep{PPIformer2024ICLR} and
GIGN~\citep{yang2023gign}; on STAG against STAG~\citep{STAG2025},
TEPCAM~\citep{TEPCAM2024}, TRAP~\citep{TRAP2025}, ConTCR~\citep{ConTCR2025},
epiTCR and epiTCR-MHC~\citep{epiTCR2023}, NetTCR-2.0~\citep{NetTCR22021} and
TITAN~\citep{Titan2021}; on PPB-Affinity against GraphDTA~\citep{GraphDTA2021},
the four TCR-pMHC predictors that expose a retrainable interface, and
DeepDTA, PIPR and D-SCRIPT carried over from the PPIRef comparison.
Every baseline was retrained from scratch on the training split of the
benchmark it is scored on, so each is fitted to the data \modelname{} sees;
PPIFormer is the single exception, used as its released PPIRef-pretrained
checkpoint.
Architectures, input representations and the full training protocol are in
Appendix~\ref{app:baseline_details}.

\paragraph{Metrics.}
We report AUC, accuracy, F1, PR-AUC, and RMSE between predicted probabilities
and labels, thresholding probabilities at $0.5$ for the binary metrics.
Baselines that output raw scores rather than calibrated probabilities (e.g.\
PPIFormer, which returns log-likelihoods) are min-max normalized first. Unless a
caption says otherwise, every entry in Tables~\ref{tab:ppi_result},
\ref{tab:ppb_result} and \ref{tab:tcr_result} is a mean $\pm$ standard
deviation over five runs, seeds $42$--$46$; what a seed varies differs by benchmark: on
PPIRef it redraws the whole $80/10/10$ partition, so the spread covers split,
initialisation and negative sampling; on STAG the test split is fixed by the
benchmark and the seed redraws the train/validation partition; on PPB-Affinity
all three splits are fixed by the benchmark and the seed varies initialisation
and batch order only. Throughout, the best result in a column is in
\textbf{bold} and the runner-up is \underline{underlined}. Split sizes are in
Table~\ref{tab:dataset_stats} and the full hyperparameter configuration in
Appendix~\ref{app:implementation_details}.

\subsection{Result Analysis}

\paragraph{PPI}
Table~\ref{tab:ppi_result} reports PPIRef results against five baselines. \modelname{} attains the highest score on all five metrics among the methods we rank; GIGN is excluded for the reason given below. Over the strongest baseline, PIPR, it gains $0.0462$ AUC and between $0.0409$ and $0.0588$ on the other four. Both methods are averaged over the same five seeds, so these margins can be read against the seed-to-seed spread of the comparison itself: each is between $6.4$ and $9.6$ times the combined standard deviation of the two methods. That margin has two parts, and only the second is architectural. DeepDTA and PIPR learn their residue embeddings from scratch while \modelname{} starts from a frozen language model, so we also report a control that keeps the language model and discards everything else: mean-pool the same cached ESM-2 embeddings, concatenate, and fit an MLP (C1 in Table~\ref{tab:ablation}). It reaches $0.9126 \pm 0.0032$. Of the $0.0462$ over PIPR, $0.0286$ is therefore the pretrained encoder and $0.0176$ is the architecture on top of it, the latter still $3.6$ times the combined standard deviation but a much smaller number than the headline margin.

Structure aware approaches show two distinct failure modes. PPIFormer~\citep{PPIformer2024ICLR} stays competitive on the ranking metrics but drops sharply on the threshold-dependent ones, a calibration effect rather than poor discrimination: its F1 ranges from $0.3635$ to $0.6994$ across seeds while its AUC moves by $\pm 0.0058$. Selecting each method's threshold on that seed's validation predictions cuts our F1 margin over it from $0.3721$ to $0.1764$, while the margins over DeepDTA and PIPR barely move (Appendix~\ref{app:threshold}). GIGN~\citep{yang2023gign} is the outlier and we flag it rather than rank by it: its AUC is the highest in the table while its accuracy and F1 vary across seeds by $\pm 0.1596$ and $\pm 0.2718$, two orders of magnitude more. A near-perfect and reproducible ordering whose operating point is close to arbitrary is not binding prediction by degree, and the likely cause is that GIGN consumes the complex interface directly and so can read the re-pairing artefact, which requires no notion of binding. We therefore exclude GIGN from the comparisons below; Appendix~\ref{app:gign} gives the evidence and what it implies for the other rows, ours included.

\paragraph{What survives a harder negative.}
All three benchmarks supply positive pairs only, so negatives must be
constructed, and re-pairing chains across complexes is the protocol PPIRef
defines. That choice leaves a cue: chain lengths alone separate the resulting
pairs to AUC $0.8138$ (Appendix~\ref{app:gign}), because the two sides of a real
interface contribute comparable numbers of residues and a uniformly re-paired
negative does not. Lacking a source of labelled non-binders we made the
re-pairing harder instead, pairing each chain with one of \emph{similar} length,
which collapses the cue to $0.5456$. Against those negatives the published
seed-$42$ model scores $0.8863$, a loss of $0.0495$ where the cue lost $0.2686$,
and still above the strongest baseline on the easy split. Since the model never
saw this distribution in training that figure is a lower bound, and length
matching bounds one cue rather than all of them.
 Against the strongest structure aware baseline (PPIFormer), \modelname{} improves AUC by $0.0968$ and F1 by $0.3721$ at the fixed cut-off ($0.1764$ under calibrated thresholds, as above); against D-SCRIPT it improves AUC by $0.2785$ and F1 by $0.6353$. Section~\ref{sec:ablation} separates the two components behind these gains, and the outcome is only partly what the design would suggest: deleting the topological descriptor costs \PHGAIN{} AUC, while deleting the structure-aware attention bias changes it by an amount we cannot distinguish from zero (A6z, $+0.0014$, $95\%$ CI $[-0.0011, +0.0038]$). We therefore leave the remaining gap to structure-aware baselines unattributed rather than assign it to a component we have not measured contributing.

\begin{table}[t]
\centering
\small
\setlength{\tabcolsep}{4pt}
\begin{tabular}{lccccc}
\toprule
\textbf{Model Name} & \textbf{AUC$\uparrow$} & \textbf{ACC$\uparrow$} & \textbf{F1$\uparrow$} & \textbf{PR-AUC$\uparrow$} & \textbf{RMSE$\downarrow$} \\
\midrule
DeepDTA & \ms{0.8691}{0.0059} & \ms{0.7785}{0.0070} & \ms{0.7791}{0.0042} & \ms{0.8765}{0.0038} & \ms{0.3865}{0.0055} \\
PIPR & \underline{\ms{0.8840}{0.0033}} & \underline{\ms{0.7944}{0.0024}} & \underline{\ms{0.7825}{0.0042}} & \underline{\ms{0.8948}{0.0026}} & \underline{\ms{0.3723}{0.0030}} \\
D-SCRIPT & \ms{0.6517}{0.0584} & \ms{0.5409}{0.0243} & \ms{0.2060}{0.1802} & \ms{0.6760}{0.0553} & \ms{0.5273}{0.0222} \\
PPIFormer & \ms{0.8334}{0.0058} & \ms{0.6546}{0.0608} & \ms{0.4692}{0.1417} & \ms{0.8692}{0.0051} & \ms{0.4636}{0.0085} \\
GIGN$^{\dagger}$ & \ms{0.9986}{0.0009} & \ms{0.7781}{0.1596} & \ms{0.6754}{0.2718} & \ms{0.9954}{0.0036} & \ms{0.3426}{0.1300} \\
\midrule

\modelname{} (ours) & \msb{0.9302}{0.0037} & \msb{0.8410}{0.0057} & \msb{0.8413}{0.0052} & \msb{0.9365}{0.0035} & \msb{0.3314}{0.0057} \\
\bottomrule
\end{tabular}
\caption{Performance on the PPIRef test set. A logistic
regression on five chain-length features alone reaches AUC $0.8138$ on this
split, so the absolute numbers include a benchmark cue no method has to learn;
Appendix~\ref{app:gign} measures it and Section~\ref{sec:ablation} reports what
survives when it is removed. $^{\dagger}$GIGN is reported for completeness but is
not a binding-prediction result and is excluded from the ranking
(Appendix~\ref{app:gign}). D-SCRIPT is the eight-member ensemble its authors'
pipeline produces (Appendix~\ref{app:dscript}).}
\label{tab:ppi_result}
\end{table}

\paragraph{TCR-pMHC}
We also evaluate on the STAG benchmark against eight TCR-pMHC predictors, all
retrained on its training split. \modelname{} leads on the threshold-dependent
metrics - accuracy $+0.0594$ and F1 $+0.0386$ over the strongest baseline,
epiTCR - and trails it on the ranking ones, by $-0.0187$ AUC and $-0.0550$
PR-AUC. We therefore make no state-of-the-art claim on TCR-pMHC: a task-specific
sequence model orders these pairs better, while \modelname{} converts its scores
into decisions more effectively under the split's $76.8\%$ class imbalance. The
table, the per-baseline reading and the protocol are in
Appendix~\ref{app:tcr}. The analysis below concerns PPIRef, the setting our
topological claim is about.

\paragraph{PPB-Affinity}
Table~\ref{tab:ppb_result} reports PPB-Affinity, which shares PPIRef's label
construction and differs from it in input size. Every baseline sits at chance on
the ranking metrics, with accuracy within a point and a half of $0.50$; the arms
that reach a non-trivial F1 do so only by collapsing towards one class in some
runs, which is what F1 standard deviations between $0.18$ and $0.37$ record
against means below $0.47$. The
unmarked rows are models built for other interaction settings and on their own
would establish only that the benchmark resists transfer. The rows marked
$^{\ast}$ remove that reading: DeepDTA, PIPR and D-SCRIPT, the general-purpose
PPI baselines of Table~\ref{tab:ppi_result}, land at chance after training on
this split itself, two of them by collapsing to a constant class rather than by
ranking badly. \modelname{} is the only method in the table that separates the
classes at all, though far below its own PPIRef performance. The limit is
therefore not transfer, and we do not attribute it to the affinity provenance of
these complexes either, because the binary label does not use it. What remains
is the input: interface regions against whole chains. The chain-length control
that separates PPIRef's pairs to $0.8138$ reaches $0.4940$ here
(Appendix~\ref{app:gign}), so the cue that makes PPIRef easy is absent, which is
the more parsimonious reading of the gap. We read PPB-Affinity as a harder
setting rather than a solved one.

Taken together, \modelname{} is the strongest method we evaluate on general PPI by a margin far outside seed variation, and is the strongest on threshold-dependent metrics but not on ranking metrics on TCR-pMHC, where a task-specific sequence model ranks better; on the whole-chain setting of PPB-Affinity it is the only method that leaves chance at all, on a benchmark where three sequence-based models trained on the split itself do not. The rest of this section concerns PPIRef, where the topological claim is made and tested.

\begin{table}[t]
\centering
\small
\setlength{\tabcolsep}{4pt}
\begin{tabular}{lccccc}
\toprule
\textbf{Model Name} & \textbf{AUC$\uparrow$} & \textbf{ACC$\uparrow$} & \textbf{F1$\uparrow$} & \textbf{PR-AUC$\uparrow$} & \textbf{RMSE$\downarrow$} \\
\midrule
ConTCR      & \ms{0.5001}{0.0017} & \ms{0.5004}{0.0016} & \ms{0.3104}{0.2898} & \ms{0.5000}{0.0026} & \ms{0.5000}{0.0000} \\
epiTCR      & \ms{0.5314}{0.0188} & \ms{0.5080}{0.0016} & \ms{0.0500}{0.0075} & \ms{0.5348}{0.0149} & \ms{0.5121}{0.0018} \\
epiTCR-MHC  & \underline{\ms{0.5342}{0.0107}} & \underline{\ms{0.5097}{0.0023}} & \ms{0.0920}{0.0058} & \underline{\ms{0.5434}{0.0088}} & \ms{0.5051}{0.0009} \\
NetTCR-2.0 & \ms{0.5048}{0.0278} & \ms{0.5028}{0.0225} & \ms{0.3341}{0.1814} & \ms{0.5024}{0.0199} & \ms{0.5003}{0.0016} \\
GraphDTA  & \ms{0.4972}{0.0165} & \ms{0.4983}{0.0057} & \underline{\ms{0.4642}{0.1988}} & \ms{0.5030}{0.0129} & \ms{0.5118}{0.0241} \\
DeepDTA$^{\ast}$ & \ms{0.5054}{0.0098} & \ms{0.5050}{0.0067} & \ms{0.3005}{0.2644} & \ms{0.5091}{0.0059} & \ms{0.5816}{0.0784} \\
PIPR$^{\ast}$ & \ms{0.5011}{0.0030} & \ms{0.5000}{0.0000} & \ms{0.2667}{0.3652} & \ms{0.4984}{0.0021} & \ms{0.5000}{0.0000} \\
D-SCRIPT$^{\ast}$ & \ms{0.5135}{0.0074} & \ms{0.5004}{0.0080} & \ms{0.0665}{0.0845} & \ms{0.5179}{0.0139} & \ms{0.6268}{0.0078} \\
\midrule
\modelname{} (ours) & \msb{0.6612}{0.0196} & \msb{0.6023}{0.0166} & \msb{0.5366}{0.0113} & \msb{0.6431}{0.0234} & \ms{0.5446}{0.0279} \\
\bottomrule
\end{tabular}
\caption{Performance on PPB-Affinity. Negatives are re-paired chains from
different complexes, as on PPIRef; the measured affinities are not thresholded
and do not enter the label.
$^{\ast}$marks the general-purpose PPI baselines of
Table~\ref{tab:ppi_result}, retrained on this benchmark's training split; the
unmarked rows are models built for other interaction settings. Every baseline
sits at chance either way. No RMSE is bolded: the column's lowest
values belong to arms that emit a constant $0.5$, which scores $0.5000$ on a
balanced split without ordering anything.}
\label{tab:ppb_result}
\end{table}

\subsection{Component Ablation}
\label{sec:ablation}

We ablate on PPIRef, the setting our topological claim concerns; STAG is a
single-family task used to test transfer, not to attribute modalities. Every arm
is a single edit to the published code path. A1--A5 edit the two terms of the fusion equation $H^{(0)} =
H_{\mathrm{seq}} + \mathbf{1}_N h_{\mathrm{topo}}^\top$, A5 degenerating both at
once as a leakage floor; A6 and A6z edit the structural attention bias; A8
sweeps the contrastive weight; A10 removes the topological descriptor and the
bias together; the labels index a larger internal grid and are therefore not
consecutive. Arms run at three seeds ($42$--$44$) and differences are tested with a paired
bootstrap over $10{,}000$ resamples on the seed-$42$ predictions, paired on
identical test rows. Table~\ref{tab:ablation} gives every arm and every metric;
Appendix~\ref{app:ablation_full} adds the per-arm reading.

\begin{table}[t]
\centering
\setlength{\tabcolsep}{2.5pt}
\footnotesize
\begin{tabular}{lccccc}
\toprule
\textbf{Arm} & \textbf{AUC$\uparrow$} & \textbf{ACC$\uparrow$} & \textbf{F1$\uparrow$} & \textbf{PR-AUC$\uparrow$} & \textbf{RMSE$\downarrow$} \\
\midrule
main\quad full model & \msb{0.9305}{0.0048} & \msb{0.8414}{0.0077} & \msb{0.8422}{0.0071} & \msb{0.9370}{0.0042} & \msb{0.3295}{0.0067} \\
\midrule
A1\quad $-$ topology & \ms{0.9149}{0.0039} & \ms{0.8246}{0.0061} & \ms{0.8208}{0.0025} & \ms{0.9248}{0.0041} & \ms{0.3516}{0.0035} \\
A1z\quad topo. zeroed & \ms{0.9165}{0.0046} & \ms{0.8276}{0.0049} & \ms{0.8263}{0.0046} & \ms{0.9259}{0.0037} & \ms{0.3460}{0.0038} \\
A1s\quad topo. shuffled & \ms{0.9157}{0.0043} & \ms{0.8266}{0.0056} & \ms{0.8240}{0.0065} & \ms{0.9255}{0.0040} & \ms{0.3473}{0.0029} \\
\midrule
A6\quad $-$ bias & \ms{0.9302}{0.0040} & \ms{0.8430}{0.0046} & \ms{0.8427}{0.0039} & \ms{0.9360}{0.0032} & \ms{0.3309}{0.0034} \\
A6z\quad bias zeroed & \ms{0.9305}{0.0060} & \ms{0.8427}{0.0072} & \ms{0.8392}{0.0068} & \ms{0.9364}{0.0050} & \ms{0.3322}{0.0076} \\
\midrule
\multicolumn{6}{l}{\emph{Sequence arms}} \\
A2\quad seq. zeroed$^{\ddagger}$ & \ms{0.8253}{0.0092} & \ms{0.7267}{0.0157} & \ms{0.7452}{0.0098} & \ms{0.8406}{0.0095} & \ms{0.4184}{0.0079} \\
A3\quad seq. $\to$ one-hot & \ms{0.9058}{0.0114} & \ms{0.8178}{0.0133} & \ms{0.8128}{0.0130} & \ms{0.9179}{0.0088} & \ms{0.3578}{0.0096} \\
A4\quad seq. $\to$ ESM-35M & \ms{0.9289}{0.0028} & \ms{0.8396}{0.0045} & \ms{0.8398}{0.0014} & \ms{0.9353}{0.0022} & \ms{0.3339}{0.0030} \\
A5\quad floor$^{\S}$ & \ms{0.5401}{0.0040} & \ms{0.5000}{0.0000} & \ms{0.6667}{0.0000} & \ms{0.5234}{0.0038} & \ms{0.6153}{0.0659} \\
\midrule
A8a\quad $-$ NT-Xent & \ms{0.9267}{0.0032} & \ms{0.8376}{0.0039} & \ms{0.8367}{0.0062} & \ms{0.9319}{0.0025} & \ms{0.3344}{0.0048} \\
A8b\quad $\alpha{=}0.25$ & \ms{0.9299}{0.0039} & \ms{0.8435}{0.0048} & \ms{0.8423}{0.0063} & \ms{0.9354}{0.0039} & \ms{0.3352}{0.0067} \\
\midrule
A10\quad $-$ topo, $-$ bias & \ms{0.9152}{0.0050} & \ms{0.8245}{0.0080} & \ms{0.8231}{0.0080} & \ms{0.9253}{0.0039} & \ms{0.3501}{0.0018} \\
\midrule
C1$^{\dagger}$\quad ESM mean-pool & \ms{0.9126}{0.0032} & \ms{0.8241}{0.0040} & \ms{0.8202}{0.0049} & \ms{0.9228}{0.0030} & \ms{0.3476}{0.0036} \\
\bottomrule
\end{tabular}
\caption{Ablation on the PPIRef test set, seeds $42$--$44$, on the
$H_0{+}H_1$ configuration of Table~\ref{tab:ppi_result}. Bold marks the full
model as the reference, not a winner. Each A arm is one flag on the released
code path. $^{\dagger}$C1 is not: the same cached embeddings mean-pooled into an
MLP, untuned, over five seeds. $^{\ddagger}$A2 and $^{\S}$A5 are not read at
face value.
Appendix~\ref{app:ablation_full} explains all three.}
\label{tab:ablation}
\end{table}

\paragraph{The topological descriptor carries protein-specific signal, not capacity.}
Deleting the topology branch (A1) costs \PHGAIN{} AUC, but on its own that number
is ambiguous: the deletion also removes $142{,}848$ parameters, so the drop is
equally consistent with the model simply being short of capacity. Two controls
separate the two explanations. A1z keeps the topology MLP and trains it but
feeds it a zero vector, so the branch reduces to a single learned bias added to
every residue of every protein: capacity intact, protein-specific information
gone. A1s changes only the data served to that branch, giving each protein
another protein's landscape, so parameter count, input distribution, descriptor
statistics and the gradient path are unchanged and only the correspondence
between a protein and its topology is destroyed. A1z recovers $0.0016$ of the
loss, so capacity accounts for about a tenth of A1's total and
protein-specific topological signal for the remaining $0.0140$. A1s agrees, the two controls landing within $0.0013$ of each other on the
seed-$42$ partition (both $p < 10^{-4}$): a wrong descriptor is worth about what
a constant one is worth, which is what a branch read for protein-specific
content predicts.

\paragraph{The sequence arms.}
Replacing the embedding with a parameter-free sinusoidal encoding of residue
index (A2$_{H_0}$) removes amino acid identity while keeping residues
distinguishable, and the model still reaches $0.9057$ AUC, above every baseline
we rank. Read against control C1 below, which keeps only the sequence pathway
and reaches $0.9126$, this says that either pathway alone gets most of the way
here, not that the structural one is privileged. One-hot identity (A3) costs a paired $-0.0216$ ($p<10^{-4}$), so residue
identity does not substitute for the pretrained representation. Scaling the
encoder four-fold slightly hurts instead of helping, a paired $-0.0037$
($p = 0.025$), so the result does not come from the size of the language model
and we keep the smallest checkpoint. A2 zeroes the embedding rather than substituting for it, which as
the table's $\ddagger$ note records also closes the coordinate pathway, so it is
a topology-only arm and the useful complement to A1: topology without sequence
(paired $-0.0999$) and sequence without topology (paired $-0.0166$) both fall
short of the full model. The floor A5 collapses to a constant, so a model with
both input channels degenerate cannot separate the pairs at all, which rules out
leakage through the splits, the caches or the metric code.
Appendix~\ref{app:ablation_full} reads each arm, reports the $H_0$-only versions
and gives the evidence behind both notes.

\paragraph{The structural attention bias does not measurably contribute.}
Without these arms the gap over structure-aware baselines could only be credited
to the attention bias by elimination, since the topological branch was the only
component measured. A6 (bias deleted with its parameters) and A6z (branch kept
and trained, its output discarded) test that directly, and neither moves AUC:
$-0.0011$ ($p = 0.44$) and $+0.0014$ ($p = 0.27$), both $95\%$ intervals
straddling zero. We therefore report the gap as unattributed. What these arms
remove is larger than ``attention bias'' suggests: the C$\alpha$ coordinates
enter the residue encoder at exactly one point, so deleting it leaves the
persistence landscape - computed from the same point cloud - as the only channel
through which 3D structure reaches the model. That direct view of the
coordinates costs nothing we can measure while the descriptor costs \PHGAIN{}
(A1), so the geometry this model uses is what the landscape summarises, not what
the dense pairwise-distance bias supplies.

\paragraph{The contrastive term contributes, but far less than the descriptor.}
Arms A8a and A8b sweep $\alpha$ in
$\mathcal{L} = \alpha\,\mathcal{L}_{\mathrm{NT\text{-}Xent}} + \beta\,\mathcal{L}_{\mathrm{BCE}}$
with everything else fixed; $\alpha = 0.5$ is the full-model row. Halving it is
indistinguishable from the published setting, and removing the term costs a
paired $-0.0054$ ($[-0.0079, -0.0028]$, $p < 10^{-4}$) - measurable, but about a
third of what the descriptor does (A1, \PHGAIN{}). The multi-objective loss is
not what carries the result.
\paragraph{What the architecture is worth.}
A10 keeps the dual-tower encoder, cross-attention and the contrastive term but
removes both structural inputs; control C1 removes the encoder as well,
mean-pooling the same cached embeddings into an MLP. Their three rows split the
$0.0179$ over C1 into two very unequal parts. The topological descriptor and the
distance bias account for $0.0153$, essentially all of it and essentially all of
A1's \PHGAIN{} on its own. The entire attention stack on top of that - four
structure-aware layers, cross-attention soft docking and the NT-Xent term -
accounts for a further $0.0026$, which is $0.4$ times the combined standard
deviation of the two arms and not a difference we can measure. On this benchmark
the descriptor is doing the work the architecture is credited with, and A6, A6z
and A8 say the same thing component by component. C1 is untuned, so $0.0179$
bounds the architectural gain from above rather than estimating it.

\section{Conclusion}
We presented \modelname{}, a dual-tower framework that encodes each protein
independently from its sequence, C$\alpha$ point cloud and persistent-homology
descriptor, and models the interaction through bidirectional cross-attention
rather than joint encoding of a bound complex. It leads every baseline we
evaluate on general PPI, leads at a fixed threshold on TCR-pMHC while being
outranked by a task-specific sequence model, and is the only method above
chance on PPB-Affinity; the dual-tower design makes per-protein
representations reusable across candidate pairs. Two results qualify the
picture: the component analysis is run on general PPI only, and the attention
machinery around the descriptor does not contribute measurably, so the margin
over structure-aware baselines is left unattributed.

\section{Limitations}
\label{sec:limitations}

\paragraph{Unbound structures as a binding prior.}
\modelname{} consumes each protein's individual structure, so the geometry it
sees is an imperfect proxy for the bound interface in strongly induced-fit
systems. The choice is deliberate, since at screening scale the bound complex
is exactly what is unavailable, but the expected cost is degradation on
flexible complexes relative to near-rigid ones, and we do not stratify by
conformational change.

\paragraph{The architecture beyond the descriptor.}
Section~\ref{sec:ablation} leaves the attention stack unsupported as a
contribution on PPIRef. The version those arms point to is a simplified model -
frozen embeddings, the topological descriptor, a small head - which we have not
trained, and whether the same holds on TCR-pMHC and on the whole-chain pairs of
PPB-Affinity, where the descriptor behaves differently, is open. The
contrastive term is also tested in a narrow regime: only interacting pairs
serve as anchors, so the labelled negatives never enter it and batches of $8$
to $16$ leave a handful of in-batch negatives per anchor.

\paragraph{Scope and descriptor design.}
Because the descriptor summarizes the whole point cloud rather than the
interface, how much of the benefit is interface-specific rather than fold-level
is unresolved. It is also fixed, non-differentiable and tied to particular
choices (Vietoris-Rips, dimensions $0$ and $1$, landscape vectorization), so it
cannot adapt to the task. Antibody-antigen and protein-peptide settings would
help characterize where the dual-tower formulation generalizes.


\clearpage


\section*{AI use statement}

Generative AI tools were used in preparing this work, and we disclose their role
as follows. Coding assistance: an AI assistant was used to write and debug
experiment infrastructure - feature-cache builders, seed drivers, aggregation
and significance-testing scripts - and to implement the ablation switches. All
such code was reviewed by the authors, and the numerical equivalence of the
batched forward path against the per-pair reference implementation is verified by
a test included in the released code. Writing assistance: an AI assistant was
used to draft and revise portions of the experimental sections. Every number
reported in this paper was produced by the runs described in
Section~\ref{sec:experiments} and checked by the authors against the run
artifacts; no result, table entry or citation was generated by an AI tool. The
research questions, experimental design, and the interpretation of the results
are the authors' own. We take responsibility for the final content of this work.

\section*{Ethics statement}

\modelname{} predicts whether two proteins bind. The intended use is to
prioritize candidates for experimental validation in target identification and
binder design, where cheap in-silico triage reduces wet-lab cost. The same
capability is dual-use: a method that ranks candidate binders cheaply could in
principle be used to search for binders to targets of concern, and the
screening throughput that motivates this work is precisely what makes such a
search less costly. We consider the marginal risk small relative to already
available tools - \modelname{} predicts a binary interaction label, not a
structure, a sequence, or a designed binder, and it has no generative component
- but it is not zero, and it grows if such predictors are placed inside a
design loop. We therefore report the method and evaluation protocol rather than
releasing tuned checkpoints for sensitive target families, and we encourage
downstream users who couple binding predictors to generative binder design to
adopt the screening practices developed for that setting. Beyond misuse, the
practical risk is over-reliance: our model is trained on the structural and
interaction distributions of PPIRef, STAG and PPB-Affinity, its predictions are not calibrated
probabilities of physical binding, and it should inform experimental
prioritization rather than substitute for it.

\section*{Reproducibility statement}

The code is uploaded as supplementary material. It is \modelname{} alone: the
architecture, the three dataset loaders, the training and evaluation entry
points, the parallel feature-cache
builders for all three benchmarks, the per-seed drivers, and the scripts that
aggregate those seeds into our row of Tables~\ref{tab:ppi_result},
\ref{tab:ppb_result} and~\ref{tab:tcr_result}. Baseline code, the ablation
switches of Section~\ref{sec:ablation}, the paired-bootstrap script and the
threshold analysis of Appendix~\ref{app:threshold} are kept out of that package
so that it runs one configuration and no flags; all are available on request.
Section~\ref{sec:experiments} gives the splits, seeds and protocol for each
benchmark; Appendix~\ref{app:Dataset_Details} gives the per-split pair counts and
Appendix~\ref{app:implementation_details} the full hyper-parameter
configuration.

Four details are easy to get wrong when re-running from source. Three are ours
and are documented in the code: the persistence descriptor requires the Rips
complex to be expanded one dimension above the homology dimension being
computed; the per-protein feature cache key carries no topological parameters,
so a cache built for one descriptor configuration must not be reused for
another; and ESM-2 accepts at most $1{,}022$ residues, so we truncate both the
embedding and the C$\alpha$ coordinates there to keep the two aligned, which on
PPB-Affinity affects $144$ of $10{,}484$ pairs and $2$ of the $1{,}034$ test
pairs (Appendix~\ref{app:implementation_details}). The fourth
is in a baseline: D-SCRIPT's contact module allocates a positional-weight buffer
for sequences of at most $2{,}000$ residues and then silently truncates, which
fails on PPB-Affinity chains of up to $3{,}302$ residues, so reproducing its row
in Table~\ref{tab:ppb_result} requires enlarging that buffer. The change leaves
the computation identical for any chain within the original limit, so it does
not affect D-SCRIPT's PPIRef numbers.


\bibliographystyle{plainnat}
\bibliography{iclr2027_conference}


\clearpage

\appendix

\section{Dataset Details}
\label{app:Dataset_Details}

PPIRef is a general PPI dataset that is not restricted to a specific interaction
family. The STAG dataset is purpose-built for TCR-pMHC interaction. All models
compared on PPB-Affinity are trained and tested on that dataset under the same
$80/10/10$ protocol used for PPIRef, so no method sees a different partition.

\paragraph{Input provenance.}
On all three benchmarks the C$\alpha$ coordinates come from experimentally
resolved structures, and on all three they are the chains of a resolved complex
rather than independently determined or predicted monomers, so the conformation
is the bound one. The granularity differs. A PPIRef row is the $6\,\text{\AA}$
interface region extracted from the complex, a median of $24$ residues per tower
and a $90$th percentile of $46$; a PPB-Affinity row is the whole chain, median
$199$ and $90$th percentile $438$; STAG supplies the chains of the TCR-pMHC
complex. This bears on the screening argument of Section~\ref{sec:intro}: an
interface region cannot be identified without the complex that defines it, so
the PPIRef numbers are obtained on an input a screening pipeline could not
construct for itself. The PPB-Affinity and STAG inputs do not have that problem,
but their conformations still come from the bound state. We state this rather
than qualify the claim away. The architecture never encodes an assembled
complex, which is the design claim; but our evaluation does not establish that
it works from unbound or predicted monomer structures, and that is the
experiment the claim needs.

We evaluate \modelname{} on three binary binding prediction benchmarks. Each pair $(P_A, P_B)$ is labeled as $1$ if the two proteins are reported to interact and $0$ otherwise. STAG supplies labelled negatives with its own split and we use them unchanged. PPIRef and PPB-Affinity supply positive pairs only, so on both we construct negatives by pairing each positive with a chain drawn from a different complex, following the re-pairing used for PPIRef in~\citet{PPIformer2024ICLR}. On PPB-Affinity the constructed pairs are written once to fixed train/val/test CSVs and reused by every method and every seed. Table~\ref{tab:dataset_stats} summarizes the resulting positive/negative counts in each split.

\begin{table}[t]
\centering
\begin{tabular}{lcccccc}
\toprule
\multirow{2}{*}{\textbf{Dataset}}
& \multicolumn{2}{c}{\textbf{Train}}
& \multicolumn{2}{c}{\textbf{Val}}
& \multicolumn{2}{c}{\textbf{Test}} \\
\cmidrule(lr){2-3}\cmidrule(lr){4-5}\cmidrule(lr){6-7}
& Pos & Neg & Pos & Neg & Pos & Neg \\
\midrule
STAG~\citep{STAG2025}                & 5{,}906  & 19{,}497 & -      & -      & 1{,}476 & 4{,}874 \\
PPIRef~\citep{PPIformer2024ICLR}     & 40{,}868 & 40{,}868 & 5{,}099 & 5{,}099 & 5{,}121 & 5{,}121 \\
PPB-Affinity~\citep{PPBAffinity2024} & 4{,}207 & 4{,}207 & 518 & 518 & 517 & 517 \\
\bottomrule
\end{tabular}
\caption{Dataset statistics for each task. STAG is released as a two-way split, so its validation fold is the $10\%$ of the train column that each seed holds out rather than a separate fold; PPIRef and PPB-Affinity both use an $80/10/10$ partition with held-out validation and test folds. The PPIRef counts are the interfaces actually used: the $80/10/10$ partition of the $51{,}755$ deduplicated keys yields $41{,}404/5{,}175/5{,}176$ entries, of which $536/76/55$ are dropped because the corresponding structure file is absent from the released \texttt{ppi\_6A} archive. The PPIRef test set therefore contains $10{,}242$ pairs. PPB-Affinity is partitioned $80/10/10$ over $10{,}484$ balanced pairs and, like PPIRef, carries an equal number of positives and negatives in every fold.}
\label{tab:dataset_stats}
\end{table}

\section{On GIGN's Near-Perfect Scores}
\label{app:gign}

Retrained on the PPIRef split, GIGN~\citep{yang2023gign} reports an AUC of
$0.9986 \pm 0.0009$ and a PR-AUC of $0.9954 \pm 0.0036$. We do not treat this as
a binding-prediction result, for three reasons that are visible in the numbers
themselves.

First, the ranking metrics and the threshold metrics disagree by two orders of
magnitude in stability. Across five seeds GIGN's AUC varies by $\pm 0.0009$
while its accuracy varies by $\pm 0.1596$ and its F1 by $\pm 0.2718$, spanning
$0.30$ to $0.98$. A model that has learned a graded notion of binding affinity
produces scores whose ordering \emph{and} whose scale are reproducible; a model
reading a near-deterministic cue produces a perfect ordering whose absolute
placement is arbitrary.

Second, the ceiling is implausible for the task. PPIRef pairs include
interactions whose strength varies over orders of magnitude and interfaces whose
geometry the bound structure does not fully determine; no method in the
literature, including those trained on far more PPI data, approaches $0.999$.
The other four baselines on this split span $0.65$--$0.88$.

Third, and most directly, the negatives invite exactly this shortcut. Following
PPIFormer's protocol, a negative pair is produced by re-pairing two chains at
random. The result is a complex that was never assembled, and assembling it
mechanically leaves geometric traces - backbone clashes, voids, spacing no
folded interface would show. GIGN takes the complex interface as its input and
is, of the baselines, the one best equipped to notice this. Distinguishing a real
interface from a mechanically re-paired one requires no model of binding.

This has a consequence beyond GIGN's own row, which we state plainly: every
number on PPIRef, ours included, is measured against these negatives, so some
share of every model's score may reflect decoy detection rather than binding
prediction. What bounds our own exposure is architectural rather than empirical.
\modelname{} encodes each protein separately and never sees the assembled
complex, so the inter-chain geometry that carries the re-pairing artefact -
backbone clashes and implausible spacing \emph{between} the two chains - is
not part of its input at all; the only cross-chain computation is attention
between two independently built residue representations. That argument limits
the channel but does not measure the residue, and none of our ablation arms
measures it either: the arm that removes both input terms (A5) also collapses
the coordinate pathway, so its being at chance says nothing about whether the
geometric artefact is readable. We therefore do not claim a bound on how much of
our PPIRef score is decoy detection, and regard a cluster-disjoint split with
harder negatives as the more informative evaluation and the main limitation of
this benchmark.

\paragraph{How much of PPIRef is available without a model.}
A trivial-feature control puts a number on the question. A logistic regression
on five chain-length features - the two lengths, their absolute difference,
their product, and the ratio of the smaller to the larger - reaches AUC
$0.8138$ on the seed-$42$ PPIRef test split, and one on the two persistence
descriptors, combined as an elementwise product and an absolute difference,
reaches $0.7905$. Neither contains any model of binding. The mechanism is that a
$6\,\text{\AA}$ interface region is mutually defined: the two sides of a real
interface contribute comparable numbers of residues, and a re-paired negative
breaks that correlation. The same length control on PPB-Affinity, whose towers
see whole chains, reaches $0.4940$.

We therefore read Table~\ref{tab:ppi_result} as follows. Of \modelname{}'s
$0.9302$, $0.8138$ is reachable from chain lengths alone; our margin over that
control is $0.1164$ and our margin over the strongest baseline is $0.0462$. The
ranking in Table~\ref{tab:ppi_result} is unaffected, since every method is
scored on the same pairs, but the absolute numbers are not a measure of binding
prediction alone, and the same caution applies to the ablation deltas of
Section~\ref{sec:ablation}, which are differences on top of that floor. One
detail is worth recording for anyone repeating this: a logistic regression on
the two feature sets \emph{concatenated} rather than combined scores exactly
$0.5000$, because the negatives re-pair the same chains and the class marginals
are identical by construction. Only interaction terms expose the cue, so a
concatenation-style control would wrongly suggest there is none.

\section{D-SCRIPT Configurations and Reproducibility}
\label{app:dscript}

D-SCRIPT is the one baseline whose reported number depends heavily on how it is
run, so Table~\ref{tab:dscript_configs} reports both configurations rather than
only the favourable one.

\begin{table}[t]
\centering
\small
\setlength{\tabcolsep}{4pt}
\begin{tabular}{lccccc}
\toprule
\textbf{Config.} & \textbf{AUC$\uparrow$} & \textbf{ACC$\uparrow$} & \textbf{F1$\uparrow$} & \textbf{PR-AUC$\uparrow$} & \textbf{RMSE$\downarrow$} \\
\midrule
Ensemble of 8 & \ms{0.6517}{0.0584} & \ms{0.5409}{0.0243} & \ms{0.2060}{0.1802} & \ms{0.6760}{0.0553} & \ms{0.5273}{0.0222} \\
Single model  & \ms{0.5158}{0.0276} & \ms{0.5130}{0.0223} & \ms{0.3345}{0.2577} & \ms{0.5571}{0.0374} & \ms{0.6125}{0.0491} \\
\bottomrule
\end{tabular}
\caption{D-SCRIPT on the PPIRef test set, means over five seeds.
Table~\ref{tab:ppi_result} carries the ensemble, the stronger of the two
configurations.}
\label{tab:dscript_configs}
\end{table}

Three observations follow. First, a single D-SCRIPT model does not learn this
task: across five seeds it sits at $0.5158 \pm 0.0276$ AUC, indistinguishable
from chance, and its F1 standard deviation ($0.2577$) is larger than its mean.
Second, ensembling eight such models recovers real signal ($0.6517 \pm 0.0584$),
so the ensemble is not merely averaging away noise but combining weak and
unstable predictors into a usable one. Third, and the reason we report a range
rather than a point, the ensemble itself is unstable across seeds: individual
ensembles span $0.5954$ to $0.7233$. On PPB-Affinity we report a three-member
ensemble per seed, and all $15$ members fall in AUROC $0.4852$--$0.5535$.

Every D-SCRIPT figure we quote is a five-seed mean rather than a best run, as
is every other number in this paper; given the spread above, a best run would
credit the baseline with a value no re-run recovers.

\section{Threshold Sensitivity and Calibration}
\label{app:threshold}

Table~\ref{tab:threshold_results} sweeps \modelname{}'s decision threshold on
PPIRef and STAG. Both peak at or near the default $0.5$: on PPIRef, F1 spans
$0.7520$--$0.8500$ and accuracy $0.7598$--$0.8501$ over thresholds $0.1$--$0.9$,
with both maxima at $0.5$; on STAG the model is almost flat, F1 spanning only
$0.7074$--$0.7203$ and accuracy $0.8619$--$0.8827$ across the same range.

Table~\ref{tab:threshold_val} asks the sharper question the main text raises: it
is not whether a model can be tuned to a good threshold, but whether the
threshold chosen on validation transfers to test. Each method's threshold is
selected on that seed's validation predictions and applied unchanged to test,
and the oracle column reports the best achievable test threshold as an upper
bound. Our F1 margin over PPIFormer falls from $0.3721$ at the fixed $0.5$
cutoff to $0.1764$ under validation-selected thresholds and $0.0845$ under the
oracle, so most of that margin was calibration rather than discrimination. The
margins over DeepDTA and PIPR barely move ($0.0622 \to 0.0590$ and
$0.0588 \to 0.0480$), because those methods are already close to calibrated.
\modelname{} itself gains little from tuning, $0.8413$ at $0.5$ against
$0.8439$ on validation-selected and $0.8458$ at the oracle, which is the
calibration claim stated as a measurement.

\begin{table}[t]
\centering
\small
\begin{tabular}{llcccc}
\toprule
\textbf{Model} & \textbf{Dataset} & \textbf{Threshold} & \textbf{Acc} & \textbf{F1} & \textbf{Predicted 1} \\
\midrule
\modelname{} & PPIRef & 0.1 & 0.7598 & 0.8026 & 7343 \\
\modelname{} & PPIRef & 0.3 & 0.8335 & 0.8466 & 5994 \\
\modelname{} & PPIRef & 0.5 & \textbf{0.8501} & \textbf{0.8500} & 5114 \\
\modelname{} & PPIRef & 0.7 & 0.8409 & 0.8268 & 4289 \\
\modelname{} & PPIRef & 0.9 & 0.7976 & 0.7520 & 3238 \\
\midrule
\modelname{} & STAG & 0.1 & 0.8619 & \textbf{0.7203} & 1659 \\
\modelname{} & STAG & 0.3 & 0.8671 & 0.7166 & 1502 \\
\modelname{} & STAG & 0.5 & 0.8706 & 0.7146 & 1404 \\
\modelname{} & STAG & 0.7 & 0.8732 & 0.7074 & 1275 \\
\modelname{} & STAG & 0.9 & \textbf{0.8827} & 0.7125 & 1115 \\
\bottomrule
\end{tabular}
\caption{\modelname{} under different classification thresholds on PPIRef and
STAG, on the seed-$42$ partition of each; the single-seed accuracy at $0.5$ is
therefore $0.8501$ rather than the five-seed $0.8410$ of
Table~\ref{tab:ppi_result}. Best value per column block in \textbf{bold}.}
\label{tab:threshold_results}
\end{table}

\begin{table}[t]
\centering
\small
\begin{tabular}{lcccc}
\toprule
\textbf{Method} & \textbf{AUC$\uparrow$} & \textbf{F1 @ $0.5$} & \textbf{F1 @ val-thr.} & \textbf{F1 @ oracle} \\
\midrule
DeepDTA   & \ms{0.8691}{0.0059} & \ms{0.7791}{0.0042} & \ms{0.7849}{0.0041} & \ms{0.7865}{0.0044} \\
PIPR      & \ms{0.8840}{0.0033} & \ms{0.7825}{0.0042} & \ms{0.7959}{0.0040} & \ms{0.7980}{0.0042} \\
PPIFormer & \ms{0.8334}{0.0058} & \ms{0.4692}{0.1417} & \ms{0.6676}{0.0411} & \ms{0.7613}{0.0067} \\
\midrule
\modelname{} (ours) & \msb{0.9302}{0.0037} & \msb{0.8413}{0.0052} & \msb{0.8439}{0.0051} & \msb{0.8458}{0.0051} \\
\bottomrule
\end{tabular}
\caption{Threshold transfer on PPIRef, five seeds. ``val-thr.'' selects the
threshold maximising F1 on that seed's validation predictions and applies it
unchanged to test; ``oracle'' is the best test threshold, an upper bound no
method can reach in practice. The gap between the last two columns is how much
a method loses to miscalibration: $0.0937$ for PPIFormer against $0.0019$ for
\modelname{}. GIGN and D-SCRIPT are omitted because they do not expose
per-seed validation predictions in the form this comparison needs.}
\label{tab:threshold_val}
\end{table}

\section{Ablation Details}
\label{app:ablation_full}

\begin{figure}[t]
    \centering
    \includegraphics[width=0.92\linewidth]{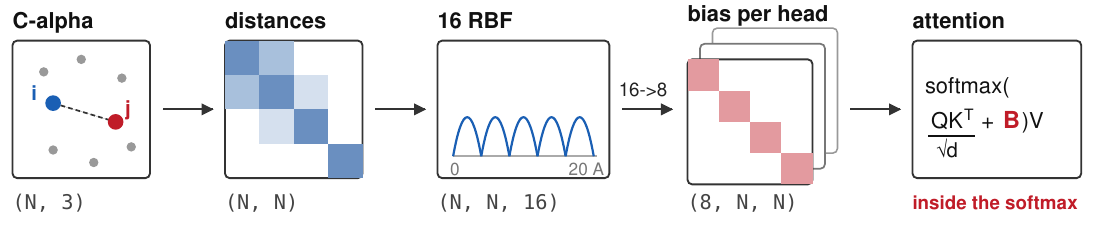}
    \caption{Construction of the structural attention bias $B_{\mathrm{struct}}$.
    Pairwise C$\alpha$ distances are expanded by $16$ radial basis functions whose
    centres are fixed on $[0,20]\,\text{\AA}$ and whose widths are learned, then
    projected to one bias matrix per attention head. The bias is added to the
    attention logits \emph{inside} the softmax, so geometry reweights attention
    rather than post-processing it, and it is dense over all $N^2$ residue pairs
    rather than restricted to a neighbourhood graph. The kernel widths and the
    $16\!\to\!8$ projection are the only learned parameters of this branch
    ($152$ per layer, $608$ across the four layers), which is the component
    removed in arm A6 of Section~\ref{sec:ablation}.}
    \label{fig:struct_bias}
\end{figure}

\begin{table}[t]
\centering
\setlength{\tabcolsep}{3pt}
\footnotesize
\begin{tabular}{llccccc}
\toprule
& \textbf{Arm} & \textbf{AUC$\uparrow$} & \textbf{ACC$\uparrow$} & \textbf{F1$\uparrow$} & \textbf{PR-AUC$\uparrow$} & \textbf{RMSE$\downarrow$} \\
\midrule
main$_{H_0}$ & Full model      & \ms{0.9283}{0.0044} & \ms{0.8407}{0.0046} & \ms{0.8387}{0.0031} & \ms{0.9352}{0.0034} & \ms{0.3322}{0.0063} \\
\midrule
A2$_{H_0}$ & seq. $\to$ sinusoidal & \ms{0.9057}{0.0045} & \ms{0.8124}{0.0055} & \ms{0.8159}{0.0046} & \ms{0.9083}{0.0046} & \ms{0.3565}{0.0049} \\
A3$_{H_0}$ & seq. $\to$ one-hot & \ms{0.9220}{0.0049} & \ms{0.8348}{0.0061} & \ms{0.8310}{0.0078} & \ms{0.9315}{0.0046} & \ms{0.3447}{0.0072} \\
A4$_{H_0}$ & seq. $\to$ ESM-35M & \ms{0.9270}{0.0040} & \ms{0.8385}{0.0071} & \ms{0.8380}{0.0063} & \ms{0.9338}{0.0033} & \ms{0.3354}{0.0053} \\
A5$_{H_0}$ & floor              & $0.4947$            & $0.5000$            & $0.6667$            & $0.4975$            & $0.5371$ \\
\bottomrule
\end{tabular}
\caption{The sequence arms on the $H_0$-only descriptor, against its own full
model. Two of them are a different ablation rather than the same one on a
different descriptor, which is why both versions are reported: A2$_{H_0}$ substitutes
sinusoidal encodings for the language-model embedding ($2{,}908{,}001$
parameters) where A2 zeroes it and keeps the projection trained
($2{,}989{,}921$, parameter-matched to the full model), and A3$_{H_0}$ projects
a $21$-d one-hot code directly ($2{,}913{,}377$) where A3 places the same code
in the leading channels of the full-width input ($2{,}989{,}921$). A4 and
A4$_{H_0}$ are the one like-for-like pair, identical at $3{,}030{,}881$
parameters, so their difference isolates the descriptor change alone.
A5$_{H_0}$'s $0.4947$ is the same constant-output artefact as A5's $0.5401$,
landing on the other side of chance.}
\label{tab:ablation_h0}
\end{table}

\paragraph{Reading the sequence arms.}
Table~\ref{tab:ablation_h0} reports the sequence arms on the $H_0$-only
descriptor. A2$_{H_0}$ keeps residues distinguishable, so the
distance-based attention bias still operates while amino acid identity is gone;
it is the largest drop outside the floor. Read against it, A3$_{H_0}$ places the value of residue identity at $0.0163$
and the value of the pretrained representation beyond identity at $0.0063$. A4
scales from the $8$M ESM-2 checkpoint to the $480$-dimensional $35$M one for no
gain on either descriptor despite four times the encoder parameters: the
three-seed means fall by $0.0013$ on $H_0$-only and $0.0017$ on $H_0{+}H_1$,
and on the seed-$42$ partition the paired differences are $-0.0019$
($p = 0.25$) and $-0.0037$ ($p = 0.025$), the latter the figure quoted in
Section~\ref{sec:ablation}. We keep the smallest checkpoint, which also keeps
the per-protein encoding cost low. A4 is the one arm whose two rows are
parameter-identical ($3{,}030{,}881$), so the pair also serves as a second
reading of the descriptor change on a matched architecture.

The two one-hot rows differ in how the code enters the tower. A3$_{H_0}$
projects a $21$-dimensional code with a correspondingly narrow input layer
($2{,}913{,}377$ parameters); A3 writes the same code into the leading channels
of the full-width input and leaves the projection at its full size
($2{,}989{,}921$, matched to the full model), so its drop is not confounded with
a smaller encoder. The two land at $0.9220 \pm 0.0049$ and
$0.9058 \pm 0.0114$ against their respective baselines, costs of $0.0063$ and
$0.0247$. We do not read that difference as an effect of either change on its
own: the arms differ in the descriptor and in the input projection at once, and
A3 carries the widest seed spread in the table ($\pm 0.0114$ against
$\pm 0.0028$ for A4 on the same three seeds). The conservative reading is that
one-hot identity is a worse substitute for the pretrained embedding than the
$H_0$-only row suggested, by an amount this pair of arms cannot localise.

\paragraph{Homology dimension.}
Because the seed also selects the data partition, the effect of a change to the
descriptor is measured on a matched partition rather than against a multi-seed
mean. On the seed-$42$ partition, replacing the $H_0$-only descriptor with the
$H_0{+}H_1$ descriptor used throughout this paper raises AUC from $0.9325$ to
$0.9358$ ($+0.0033$), PR-AUC from $0.9377$ to $0.9418$, and F1 from $0.8469$ to
$0.8500$, while lowering RMSE from $0.3302$ to $0.3224$. All five metrics
improve, though the margin is comparable to the seed-to-seed standard deviation
in Table~\ref{tab:ppi_result}, so we read this as evidence that $H_1$ does not
hurt rather than as a large effect.

\paragraph{Why A2$_{H_0}$ uses a sinusoidal encoding, and what A2 measures instead.}
The natural reading of ``no sequence'' is to replace $H_{\mathrm{seq}}$ with a
single constant, but that makes every token identical, and attention over
identical values returns that value regardless of the additive structural bias:
the model becomes blind to geometry. Measured directly, the output moved by
$0.000000$ under a fivefold perturbation of the input coordinates under that
configuration, so it measures nothing about the sequence pathway.
A2$_{H_0}$ therefore uses a parameter-free sinusoidal encoding of residue index,
which removes amino acid identity while keeping residues distinguishable so that
the distance-based bias remains active.

A2 is the constant-substitution arm that argument warns against, and we report
it because the configuration is still informative once it is read correctly:
unlike A5 its topology branch is live, so its output does vary across pairs
($10{,}237$ distinct values over $10{,}242$ test pairs on seed $42$, against
A5's two) and its $0.8253 \pm 0.0092$ is a real measurement of a real, if
narrower, model. We
verified the degeneracy on the trained checkpoint rather than assuming it:
scaling every C$\alpha$ coordinate by a factor of five leaves the predicted
probability unchanged to all printed digits on every pair we tested, while the
same perturbation moves A3 by up to $1.1 \times 10^{-3}$ and A1z by up to
$6.9 \times 10^{-3}$. What survives in A2 is the persistence landscape, which is
computed from the point cloud outside the attention stack and so is unaffected.
Nothing else protein-specific survives, since the encoder sees geometry only
through the distance bias of Section~\ref{sec:method}. A2's score is therefore a
topology-only reading and belongs beside A1 - which deletes topology and keeps sequence and
geometry - rather than beside A2$_{H_0}$. A5, the floor arm, degenerates both
terms and is a leakage check rather than a structure-only arm. It behaves as the
argument above predicts: with both $H_{\mathrm{seq}}$ and
$h_{\mathrm{topo}}$ constant the network has no protein-specific input left, and
its output takes two adjacent float values over the whole test set, varying by $1.0$--$1.4 \times
10^{-7}$ across the three seeds - two units in the last place of
\texttt{float32} at that magnitude, and a single number once rounded to six
decimals. Its accuracy of $0.5000$ and F1 of $0.6667$ follow from that constant
sitting above the $0.5$ cut-off on a balanced split. We flag the nominal AUROC
of $0.5401 \pm 0.0040$ rather than report it as a result: a ranking metric on a
constant score is decided entirely by how the tie-breaking falls out of
last-bit arithmetic, and the $H_0$-only row's $0.4947$ is the same artefact
landing on the other side of chance. Neither number measures the model. What
the arm establishes is the constancy itself, which is what a leakage check
needs.

\section{Implementation Details}
\label{app:implementation_details}

\paragraph{The NT-Xent objective in full.}
With $\mathcal{P} = \{i : y_i = 1\}$ the positive pairs of a mini-batch and
$\mathrm{sim}_{ij} = \langle z_A^{(i)}, z_B^{(j)}\rangle / \tau$ the similarity
between $\ell_2$-normalised pooled embeddings,
\[
\mathcal{L}_{\mathrm{NT\text{-}Xent}}
= \frac{1}{2 |\mathcal{P}|} \sum_{i \in \mathcal{P}}
\Bigg[
- \log \frac{\exp(\mathrm{sim}_{ii})}{\sum_{j} \exp(\mathrm{sim}_{ij})}
- \log \frac{\exp(\mathrm{sim}_{ii})}{\sum_{j} \exp(\mathrm{sim}_{ji})}
\Bigg].
\]
Each positive is its own row's and column's only positive; every other index in
the batch acts as an in-batch negative. Note that the benchmark's labelled
negatives never serve as anchors, which Section~\ref{sec:limitations} takes up.

We implement \modelname{} in PyTorch~2.6 with HuggingFace Transformers~5.3, GUDHI~3.11, and BioPython~1.81. All experiments are conducted on a single NVIDIA CUDA GPU. The complete configuration is summarized in Table~\ref{tab:hyperparameters}.

One input limit is ours and we state it alongside the baseline limit of the reproducibility statement: ESM-2 accepts at most $1{,}022$ residues, and we truncate the C$\alpha$ coordinates to match so the two modalities stay aligned. No PPIRef or STAG chain reaches that length. On PPB-Affinity $152$ of $20{,}968$ chain slots do, affecting $144$ of $10{,}484$ pairs and $2$ of the $1{,}034$ test pairs, so the effect on Table~\ref{tab:ppb_result} is negligible, but a longer-chain benchmark would need a windowed encoder.

\begin{table}[t]
\centering
\small
\begin{tabular}{ll}
\toprule
\textbf{Hyperparameter} & \textbf{Value} \\
\midrule
\multicolumn{2}{l}{\textit{Sequence features}} \\
ESM-2 model                                  & \texttt{facebook/esm2\_t6\_8M\_UR50D} \\
ESM hidden dimension $d_{\text{esm}}$        & $320$ \\
\midrule
\multicolumn{2}{l}{\textit{Topological features}} \\
Filtration                                   & Vietoris-Rips \\
Max edge length                              & $20\,\text{\AA}$ \\
Max homology dimension                       & $1$ ($H_0+H_1$) \\
Number of landscapes $n_{\text{land}}$       & $3$ \\
Resolution $n_{\text{pts}}$                  & $50$ \\
\midrule
\multicolumn{2}{l}{\textit{Architecture}} \\
Model dimension $d_{\text{model}}$           & $256$ \\
Graph Transformer layers $L$                 & $4$ \\
Attention heads $H$                          & $8$ \\
FFN hidden dimension                         & $512$ \\
Dropout                                      & $0.1$ \\
RBF kernels (structural bias)                & $16$ (cutoff $20\,\text{\AA}$) \\
Cross-attention heads                        & $8$ \\
Activation                                   & GELU \\
Normalization                                & Pre-LayerNorm \\
\midrule
\multicolumn{2}{l}{\textit{Loss}} \\
$\alpha$ (contrastive)                       & $0.5$ \\
$\beta$ (BCE)                                & $0.5$ \\
NT-Xent temperature $\tau$                   & $0.07$ \\
\midrule
\multicolumn{2}{l}{\textit{Optimization}} \\
Optimizer                                    & AdamW \\
Learning rate                                & $1\!\times\!10^{-4}$ \\
Weight decay                                 & $1\!\times\!10^{-5}$ \\
Gradient clipping (max-norm)                 & $1.0$ \\
Batch size                                   & $32$ on PPIRef, $16$ on STAG and PPB-Affinity \\
Maximum epochs                               & $50$ on PPIRef, $100$ on STAG and PPB-Affinity \\
Early-stopping patience                      & $10$ on PPIRef, $20$ on STAG and PPB-Affinity \\
Validation split (when held-out absent)      & $10\%$ \\
\bottomrule
\end{tabular}
\caption{\modelname{} hyperparameters used across all experiments unless otherwise stated.}
\label{tab:hyperparameters}
\end{table}

\section{Screening Throughput}
\label{app:runtime}

The dual-tower design is motivated by screening cost, so we measure it rather
than assert it. All numbers are single-GPU (one NVIDIA A6000), batch size one,
on a $60$-chain sample of PPB-Affinity whole chains (mean $191$ residues; the
benchmark's own mean is $242$), which is the input a screening pipeline would
actually hold. Table~\ref{tab:runtime} gives the
per-stage cost.

\begin{table}[h]
\centering
\small
\begin{tabular}{lcc}
\toprule
\textbf{Stage} & \textbf{Cost} & \textbf{Recomputed per candidate pair?} \\
\midrule
Persistence landscape & $101$ ms / protein (median $49$) & no \\
Tower encoding (Phase 1) & $2.3$ ms / protein & no \\
Fusion $+$ head (Phase 2) & $0.78$ ms / pair & yes \\
\bottomrule
\end{tabular}
\caption{Per-stage cost. Only the last row scales with the number of candidates.}
\label{tab:runtime}
\end{table}

Scoring one fixed target against $10{,}000$ candidates therefore costs $17$
minutes end to end from structures, of which $16.9$ minutes is the one-off
persistence computation over the candidate library and $0.5$ minutes is tower
encoding plus fusion. Re-screening a second target against the same library
recomputes only the last row of Table~\ref{tab:runtime}, $8$ seconds for
$10{,}000$ pairs, provided the tower representations are kept; recomputing them
as well returns the full $0.5$ minutes. Co-folding the same $10{,}000$ pairs
at the minutes-per-pair rate cited in Section~\ref{sec:intro} is on the order of
$10^3$ GPU-hours.

The persistence step is both the dominant term and the one that scales worst.
Across the chains we timed it ranges from a few milliseconds to $839$ ms, and
the slowest cases are the longest: $814$ residues costs $0.84$ s against $49$ ms
at the median. The Vietoris-Rips complex is expanded one dimension above the
homology dimension being computed, so the simplex count grows quickly with both
chain length and the $20\,\text{\AA}$ cutoff. For a library that is fixed across
many targets this is amortised and the descriptor can be shipped alongside the
structures; for a library that changes every query it would become the
bottleneck, and an alpha complex would be the natural substitute.

\begin{figure}[t]
    \centering
    \includegraphics[width=\linewidth]{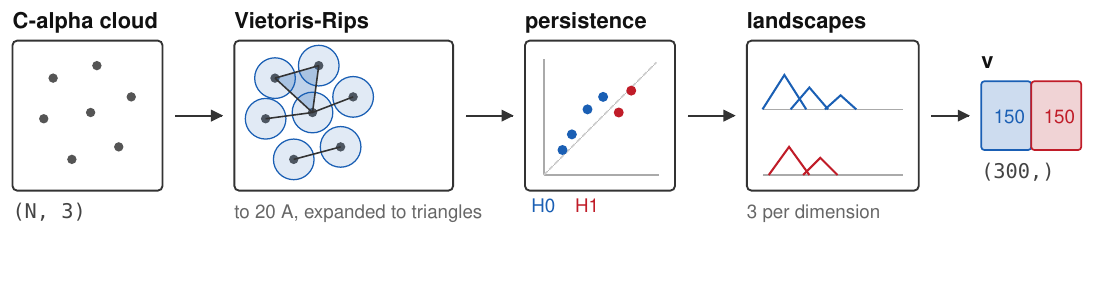}
    \caption{Construction of the topological descriptor. A Vietoris-Rips
    filtration on the C$\alpha$ cloud, expanded one dimension above the homology
    dimension so that triangles are present, gives $H_0$ and $H_1$ persistence;
    each is summarised by three landscapes on a $50$-point grid and the two
    blocks are concatenated into one $300$-dimensional vector.}
    \label{fig:Method_Topo}
\end{figure}

\section{Persistent Homology Primer}
\label{app:tda_primer}

We provide a self-contained summary of the topological constructions used by \modelname{}. The exposition follows the persistent homology pipeline employed in the topology branch of our model: \emph{point cloud $\to$ simplicial complex $\to$ filtration $\to$ persistence diagram $\to$ persistence landscape}.

\paragraph{Persistent homology.}
Persistent homology studies how topological features of a dataset evolve across multiple scales. Given a point cloud or, more generally, a filtered space, it provides a multi-scale summary of topological invariants, such as connected components ($H_0$), loops ($H_1$), and higher-dimensional voids ($H_k$ for $k \geq 2$), by tracking when these features appear and disappear as a scale parameter varies. Unlike local geometric statistics, this multi-scale summary is invariant under translations and rotations of the input and is provably stable: small perturbations of the input cause only small changes in the resulting summary~\citep{CohenSteiner2007Stability}. Its construction relies on four key components, simplicial complexes, filtrations, persistence diagrams, and functional summaries such as persistence landscapes~\citep{bubenik2015statistical}, which we describe in turn below.

\paragraph{Simplicial complex.}
A simplicial complex is a combinatorial object used to encode the topology of a dataset. Formally, a finite abstract simplicial complex $K$ over a vertex set $V$ is a collection of subsets $\sigma \subseteq V$ such that if $\sigma \in K$, then every subset $\tau \subseteq \sigma$ also belongs to $K$. Each $\sigma$ is called a simplex, and its dimension is $\dim(\sigma) = |\sigma| - 1$. In applications, simplicial complexes are often constructed from data points $X \subset \mathbb{R}^d$, for example via the Vietoris-Rips complex~\citep{Vietoris1927,EdelsbrunnerHarer2010},
\[
\mathrm{VR}_\epsilon(X)
= \left\{ \sigma \subseteq X \;\middle|\; \|x - y\| \le \epsilon \quad \forall\, x, y \in \sigma \right\},
\]
which connects all subsets of points whose pairwise distances are below the scale $\epsilon$.

\paragraph{Homology dimension versus simplex dimension.}
These two indices are distinct and are a common source of confusion, so we state
the relationship explicitly. A simplex $\sigma$ has dimension $|\sigma| - 1$: a
vertex is a $0$-simplex, an edge a $1$-simplex, a triangle a $2$-simplex. The
homology group $H_k$ is computed from the chain complex
$\cdots \to C_{k+1} \to C_k \to C_{k-1} \to \cdots$, where $C_k$ is generated by
the $k$-simplices of $K$, and $H_k = \ker \partial_k / \operatorname{im} \partial_{k+1}$.
Because $H_k$ quotients by the image of $\partial_{k+1}$, a complex that contains
no simplices of dimension $k+1$ has $\operatorname{im} \partial_{k+1} = 0$ and
therefore reports every $k$-cycle as a feature that never dies. Concretely,
$H_1$ counts loops that are not filled in by triangles, so a Vietoris-Rips
complex expanded only to dimension $1$ (edges) has no triangles with which to
fill any loop and yields no finite $H_1$ intervals at all. Computing $H_k$
therefore requires expanding the complex to dimension $k+1$: \modelname{} uses
$H_0$ and $H_1$, and expands the Vietoris-Rips complex to dimension $2$.

\paragraph{Filtration.}
A filtration is a nested sequence of simplicial complexes indexed by a scale parameter,
\[
K_{\epsilon_1} \subseteq K_{\epsilon_2} \subseteq \cdots \subseteq K_{\epsilon_T},
\qquad
\epsilon_1 \le \epsilon_2 \le \cdots \le \epsilon_T.
\]
More generally, a filtration can be induced by a function $f: K \to \mathbb{R}$, where each sublevel set $K_\alpha = \{\sigma \in K \mid f(\sigma) \le \alpha\}$ forms a simplicial complex. Persistent homology studies how topological features evolve along this filtration.

\paragraph{Persistence diagram.}
For each dimension $k$, applying homology to a filtration yields a sequence of homology groups $H_k(K_\epsilon)$ and tracks the birth and death of homological features as $\epsilon$ varies. Each feature is associated with a persistence interval $(b_i, d_i)$, where $b_i$ and $d_i$ denote the parameter values at which the feature appears and disappears, respectively. The persistence diagram is defined as the multiset
\[
D_k = \{(b_i, d_i) \in \mathbb{R}^2 \mid i = 1, \dots, n_k\}.
\]
Each point represents a topological feature, and its persistence $d_i - b_i$ reflects its scale stability: features with larger persistence are typically interpreted as more salient.

\paragraph{Persistence landscape.}
To enable statistical analysis and compatibility with machine learning pipelines, persistence diagrams are often mapped to functional representations. The persistence landscape~\citep{PersistenceLandscape2020} is a sequence of functions $\lambda_k : \mathbb{R} \to \mathbb{R}$ defined as follows. For each interval $(b_i, d_i)$, define a triangular function
\[
\Lambda_i(t) = \max\!\big(0,\; \min(t - b_i,\; d_i - t)\big).
\]
The $k$-th persistence landscape is then the $k$-th largest value of $\{\Lambda_i(t)\}_i$ at each $t$,
\[
\lambda_k(t) = k\text{-th largest value of } \{\Lambda_i(t)\}_i.
\]
The collection $\{\lambda_k(t)\}_{k \ge 1}$ provides a stable, vectorizable representation of the persistence diagram and can be discretized on a fixed grid to yield a finite dimensional feature vector. \modelname{} uses persistence landscapes derived from the $H_0$ and $H_1$ persistence diagrams of a Vietoris-Rips filtration~\citep{Vietoris1927,EdelsbrunnerHarer2010} on the C$\alpha$ point cloud as a global topological descriptor of each protein.

\section{TCR-pMHC Results on STAG}
\label{app:tcr}

Section~\ref{sec:experiments} summarises this benchmark in one paragraph and
leaves the analysis here, because the topological claim the paper argues is
about general PPI: STAG is a single-family task in which the fold is largely
shared across pairs, and we use it to test whether the architecture transfers,
not to attribute modalities.

\begin{table}[t]
\centering
\small
\setlength{\tabcolsep}{4pt}
\begin{tabular}{lccccc}
\toprule
\textbf{Model Name} & \textbf{AUC$\uparrow$} & \textbf{ACC$\uparrow$} & \textbf{F1$\uparrow$} & \textbf{PR-AUC$\uparrow$} & \textbf{RMSE$\downarrow$} \\
\midrule
STAG & \ms{0.8564}{0.0077} & \ms{0.7671}{0.0073} & \ms{0.6106}{0.0065} & \ms{0.6801}{0.0154} & \ms{0.3907}{0.0059} \\
TEPCAM & \ms{0.6634}{0.0618} & \ms{0.7248}{0.0856} & \ms{0.2939}{0.0396} & \ms{0.3794}{0.0449} & \ms{0.4531}{0.0443} \\
TRAP & \ms{0.6705}{0.0956} & \ms{0.7721}{0.0105} & \ms{0.1668}{0.0988} & \ms{0.4089}{0.0996} & \ms{0.4403}{0.0240} \\
ConTCR & \ms{0.7033}{0.0030} & \ms{0.6358}{0.0174} & \ms{0.4644}{0.0037} & \ms{0.3982}{0.0084} & \ms{0.4655}{0.0019} \\
epiTCR & \msb{0.9031}{0.0025} & \underline{\ms{0.8120}{0.0034}} & \underline{\ms{0.6766}{0.0039}} & \msb{0.8187}{0.0029} & \underline{\ms{0.3692}{0.0010}} \\
epiTCR-MHC & \underline{\ms{0.8945}{0.0020}} & \ms{0.7869}{0.0038} & \ms{0.6468}{0.0044} & \ms{0.7412}{0.0032} & \ms{0.3866}{0.0011} \\
NetTCR-2.0 & \ms{0.8003}{0.0162} & \ms{0.8051}{0.0118} & \ms{0.5172}{0.0600} & \ms{0.5858}{0.0374} & \ms{0.3754}{0.0067} \\
TITAN & \ms{0.6751}{0.0004} & \ms{0.7691}{0.0000} & \ms{0.0000}{0.0000} & \ms{0.3640}{0.0010} & \ms{0.4075}{0.0000} \\
\midrule
\modelname{} (ours) & \ms{0.8844}{0.0032} & \msb{0.8714}{0.0037} & \msb{0.7152}{0.0085} & \underline{\ms{0.7637}{0.0101}} & \msb{0.3319}{0.0052} \\
\bottomrule
\end{tabular}
\caption{Performance on the TCR-pMHC benchmark, STAG test split.}
\label{tab:tcr_result}
\end{table}

\paragraph{Reading the table.}
Table~\ref{tab:tcr_result} compares \modelname{} against TCR-pMHC predictors on the STAG benchmark. The strongest baseline is epiTCR (AUC $0.9031 \pm 0.0025$, PR-AUC $0.8187 \pm 0.0029$), with its MHC-aware variant close behind and STAG itself at $0.8564 \pm 0.0077$. TRAP and TITAN reach majority-class accuracy with F1 far below it ($0.1668$ and $0.0000$), so their predictions collapse towards one class - TITAN identically in all five runs, TRAP with an F1 standard deviation exceeding its mean. TEPCAM and ConTCR stay below $0.71$ AUC. This may reflect distributional differences from these methods' original training data, though preprocessing and calibration may also contribute.

\modelname{} leads on the threshold-dependent metrics and trails on the ranking ones, and we state both directions against the combined standard deviation of the two methods being compared. It is ahead on accuracy ($+0.0594$ over epiTCR, $11.8\times$), F1 ($+0.0386$, $4.1\times$) and RMSE ($-0.0373$, $7.0\times$). It is behind epiTCR on AUC ($-0.0187$, $4.6\times$) and PR-AUC ($-0.0550$, $5.2\times$), by margins as far outside seed variation as our own leads. On this benchmark, therefore, \modelname{} is not the best-ranking model: a sequence-only TCR-pMHC predictor orders pairs better, while \modelname{} converts its scores into decisions more effectively under the $76.8\%$ class imbalance of this split. We do not claim state of the art on TCR-pMHC, and read the result as evidence that a general-purpose PPI architecture does not automatically dominate a task-specific one on that task's own benchmark.

\section{Baseline Model Details}
\label{app:baseline_details}

We summarize each baseline by its architecture family and the input modalities it consumes. The \emph{Architecture} column reports the dominant building block (e.g., 1D CNN, Transformer, GNN, contrastive learning), and the \emph{Input Repr.} column describes which protein representations are fed into the model.

\paragraph{PPI baselines.}
PPI baselines take two protein representations as input and produce a binding score. DeepDTA was originally proposed for drug-target affinity and consumes 1D-tokenized SMILES and protein sequences; we adapt it to PPI by feeding two protein sequences in place of the drug and target inputs. PIPR is a sequence-only PPI predictor with a Siamese residual RCNN. D-SCRIPT couples a pretrained Bepler-Berger language model with a CNN that predicts an inter-chain contact map and pools it into a binding probability. PPIFormer operates on the 3D structure of the interface using an SE(3)-equivariant Equiformer encoder pretrained with masked modeling on PPIRef. Table~\ref{tab:ppi_baselines} summarizes these baselines.

\paragraph{Training protocol per baseline.}
DeepDTA, PIPR, D-SCRIPT and GIGN were each retrained from scratch on the PPIRef
split used by \modelname{}, so all four are evaluated under the same training
distribution and the same train/val/test partition. GIGN was designed for
protein-ligand binding affinity; retraining it here adapts a geometric
interaction network to a different interaction type, and
Appendix~\ref{app:gign} reads its scores accordingly. PPIFormer is already
trained on PPIRef in its self-supervised masked-modeling form, so we use the
released checkpoint without further training, and it is the one exception in
Tables~\ref{tab:ppi_result} and~\ref{tab:ppb_result}. This separation reflects what
is reproducibly available for each method rather than a methodological
preference.

\paragraph{The general-purpose PPI baselines on PPB-Affinity.}
DeepDTA, PIPR and D-SCRIPT are carried over to PPB-Affinity and retrained on its
training split. Two details differ from their PPIRef runs, both forced by the
benchmark. First, our PPB-Affinity negatives are constructed once and written to fixed
train/val/test CSVs rather than drawn per run, so a seed varies initialisation
and batch order only; this is also how \modelname{}'s PPB seeds
vary, and the error bars in Table~\ref{tab:ppb_result} are comparable for that
reason. Second, the sequence truncation length rises from $100$ to $512$. The
shorter value was chosen for PPIRef, whose rows are extracted interface
fragments; PPB-Affinity rows are whole chains with a median length of $199$
residues and a $95$th percentile near $500$, so keeping $100$ would discard most
of every chain and understate the baselines. D-SCRIPT additionally drops pairs
in which either chain is shorter than its contact CNN's kernel width, which
removes $82$, $20$ and $18$ pairs from train, validation and test respectively;
the same floor applies to its PPIRef runs.

On STAG the situation is uniform: every baseline in Table~\ref{tab:tcr_result}
was retrained from scratch on the STAG training split rather than evaluated from
a released checkpoint. This matters for reading that table, because several of
these predictors are much stronger here than a transfer evaluation of the same
architectures would suggest - epiTCR, for instance, reaches an AUC of $0.9031$
after retraining. Reported numbers for TCR-pMHC predictors are therefore not
comparable across papers unless the training protocol is matched.

\begin{table}[t]
\centering
\small
\begin{tabular}{llll}
\toprule
\textbf{Model} & \textbf{Year} & \textbf{Architecture} & \textbf{Input Repr.} \\
\midrule
DeepDTA~\citep{DeepDTA2018}         & 2018 & CNN                                 & Two protein sequences \\
PIPR~\citep{PIPR2019}               & 2019 & RCNN                  & Two protein sequences \\
D-SCRIPT~\citep{DSCRIPT2021}        & 2021 & Pretrained PLM + CNN     & Two protein sequences \\
PPIFormer~\citep{PPIformer2024ICLR} & 2024 & Transformer  & 3D structure \\
GIGN~\citep{yang2023gign}           & 2023 & Geometric Interaction GNN           & 3D structure \\
GraphDTA~\citep{GraphDTA2021}       & 2021 & GNN + CNN                           & Two protein sequences \\
\bottomrule
\end{tabular}
\caption{Summary of the PPI baselines. GraphDTA is evaluated on PPB-Affinity only;
the others are evaluated on PPIRef and, where marked in Table~\ref{tab:ppb_result},
retrained on PPB-Affinity as well.}
\label{tab:ppi_baselines}
\end{table}

\paragraph{TCR-pMHC baselines.}
TCR-pMHC baselines vary in which components of the TCR-pMHC system they consume: some use only CDR3$\beta$ and peptide, while others additionally use CDR3$\alpha$ or MHC information; CDR3$\alpha$ and CDR3$\beta$ are the two hypervariable loops of the TCR. STAG is the only structure-based method in this group and operates directly on the 3D complex. Sequence-based methods differ in encoder choice (1D CNN, Transformer, language model, random forest) and in whether they incorporate contrastive learning to improve generalization. Table~\ref{tab:tcr_baseline_architectures} lists the architectures and Table~\ref{tab:tcr_baseline_inputs} the input representations.

\begin{table}[t]
\centering
\small
\begin{tabular}{lll}
\toprule
\textbf{Model} & \textbf{Year} & \textbf{Architecture} \\
\midrule
STAG~\citep{STAG2025}              & 2025 & Graph Neural Network \\
TEPCAM~\citep{TEPCAM2024}          & 2024 & CNN + attention \\
TRAP~\citep{TRAP2025}              & 2025 & Transformer + contrastive learning \\
ConTCR~\citep{ConTCR2025}          & 2025 & Cross-attention Transformer + contrastive learning \\
epiTCR~\citep{epiTCR2023}          & 2023 & Random Forest \\
epiTCR-MHC~\citep{epiTCR2023}      & 2023 & Random Forest (MHC-aware) \\
NetTCR-2.0~\citep{NetTCR22021}     & 2021 & CNN \\
TITAN~\citep{Titan2021}            & 2021 & Bimodal attention \\
\bottomrule
\end{tabular}
\caption{Summary of TCR-pMHC baseline model architectures.}
\label{tab:tcr_baseline_architectures}
\end{table}

\begin{table}[t]
\centering
\small
\setlength{\tabcolsep}{5pt}
\begin{tabular}{lcccccccc}
\toprule
\multirow{2}{*}{\textbf{Model}} 
& \multicolumn{4}{c}{\textbf{Sequence}} 
& \multicolumn{4}{c}{\textbf{Structure}} \\
\cmidrule(lr){2-5}\cmidrule(lr){6-9}
& \textbf{Peptide} & \textbf{MHC} & \textbf{CDR3$\alpha$} & \textbf{CDR3$\beta$}
& \textbf{Peptide} & \textbf{MHC} & \textbf{CDR3$\alpha$} & \textbf{CDR3$\beta$} \\
\midrule
STAG          
&  &  &  &  
& \checkmark & \checkmark & \checkmark & \checkmark \\

TEPCAM      
& \checkmark &  &  & \checkmark 
&  &  &  &  \\

TRAP          
& \checkmark &  &  & \checkmark 
& \checkmark$^{\ddagger}$ & \checkmark$^{\ddagger}$ &  &  \\

ConTCR      
& \checkmark & \checkmark &  & \checkmark 
&  &  &  &  \\

epiTCR      
& \checkmark &  &  & \checkmark 
&  &  &  &  \\

epiTCR-MHC
& \checkmark & \checkmark &  & \checkmark
&  &  &  &  \\

NetTCR-2.0 
& \checkmark &  & \checkmark & \checkmark 
&  &  &  &  \\

TITAN        
& \checkmark$^{\dagger}$ &  &  & \checkmark 
&  &  &  &  \\
\bottomrule
\end{tabular}
\caption{Input representations used by TCR-pMHC baseline models. $^{\dagger}$TITAN represents epitopes using SMILES rather than standard amino-acid sequence encoding. $^{\ddagger}$TRAP is the one baseline whose structural input is predicted rather than resolved: it folds each pMHC with AlphaFold-Multimer and extracts features from the epitope's local environment, so the MHC enters through that structure and not as a sequence embedding.}
\label{tab:tcr_baseline_inputs}
\end{table}

\end{document}